\documentclass[conference]{IEEEtran}
\IEEEoverridecommandlockouts

\usepackage{cite}
\usepackage{amsmath,amssymb,amsfonts}
\usepackage{algorithmic}
\usepackage{graphicx}
\usepackage{textcomp}
\usepackage{xcolor}
\usepackage{bm}
\usepackage{gensymb}
\usepackage{enumitem}
\usepackage{url}
\def\BibTeX{{\rm B\kern-.05em{\sc i\kern-.025em b}\kern-.08em
    T\kern-.1667em\lower.7ex\hbox{E}\kern-.125emX}}
\begin{document}

\title{Docking and Persistent Operations for a Resident Underwater Vehicle\\
}

\author{\IEEEauthorblockN{Leonard Günzel, Gabrielė Kasparavičiūtė, Ambjørn Grimsrud Waldum, \\ Bjørn-Magnus Moslått, Abubakar Aliyu Badawi, Celil Yılmaz,\\ Md Shamin Yeasher Yousha, Robert Staven, Martin Ludvigsen}
\IEEEauthorblockA{\textit{Applied Underwater Robotics Lab} \\
\textit{NTNU}\\
Trondheim, Norway \\
leonard.gunzel@ntnu.no}}

\maketitle
\begin{abstract}
Our understanding of the oceans remains limited by sparse and infrequent observations, primarily because current methods are constrained by the high cost and logistical effort of underwater monitoring, relying either on sporadic surveys across broad areas or on long-term measurements at fixed locations.
To overcome these limitations, monitoring systems must enable persistent and autonomous operations without the need for continuous surface support.
Despite recent advances, resident underwater vehicles remain uncommon due to persistent challenges in autonomy, robotic resilience, and mechanical robustness, particularly under long-term deployment in harsh and remote environments.
This work addresses these problems by presenting the development, deployment, and operation of a resident infrastructure using a docking station with a mini-class Remotely Operated Vehicle (ROV) at 90\,m depth. The ROV is equipped with enhanced onboard processing and perception, allowing it to autonomously navigate using USBL signals, dock via ArUco marker-based visual localisation fused through an Extended Kalman Filter, and carry out local inspection routines.
The system demonstrated a 90\,\% autonomous docking success rate and completed full inspection missions within four minutes, validating the integration of acoustic and visual navigation in real-world conditions.
These results show that reliable, untethered operations at depth are feasible, highlighting the potential of resident ROV systems for scalable, cost-effective underwater monitoring.
\end{abstract}


\section{Introduction}
As economic activity increasingly expands from terrestrial to underwater domains \cite{scholaert2020blue}, the need for more frequent monitoring and inspection of subsea infrastructure and its surrounding environment is growing rapidly. Today, such inspections are typically conducted using sonar-equipped autonomous underwater vehicles (AUVs) and survey vessels for broad-area scans, or ROVs for manual high-detail inspections of complex structures \cite{nauert2023inspection}.

These operations are resource-intensive, requiring significant time due to mobilisation, transit and deployment of the ROV, costs in terms of crew and vessel usage, and logistical coordination. Further, in light of recent sabotage events targeting critical underwater infrastructure, the need for continuous local presence has become increasingly evident \cite{soldi2023monitoring}.

The limitations of time and spatial coverage affect not only infrastructure monitoring but also environmental observations. In remote marine areas, our understanding of the ocean is often built on sparse observations, extrapolated from either short-term, spatially distributed campaigns such as infrequent AUV deployments, or long-term, spatially fixed installations like seafloor observatories \cite{whitt2020future}. 

\begin{figure}[!htbp]
\centering
\includegraphics[width=0.45\textwidth]{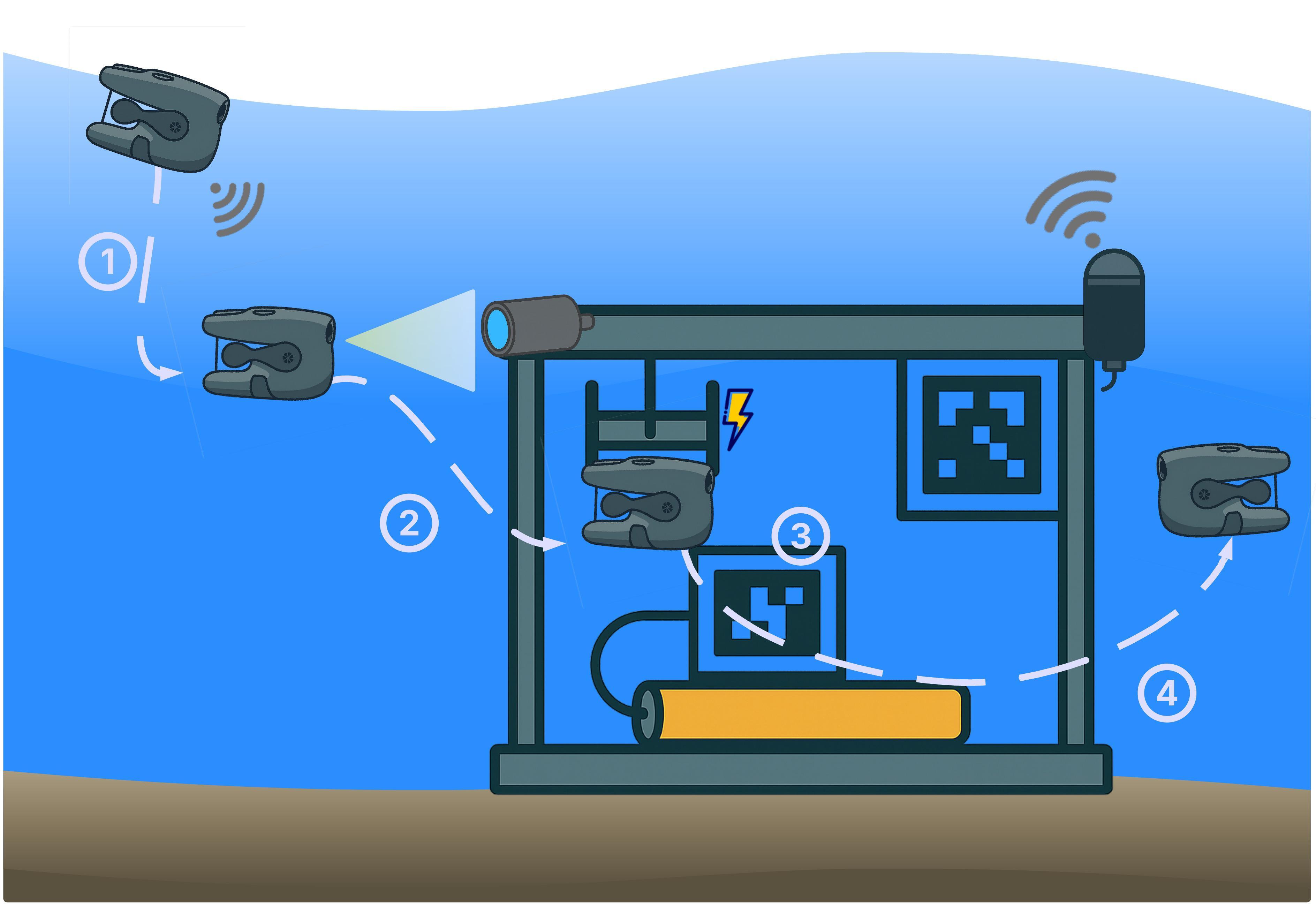}
\caption{Autonomous homing, docking and inspection workflow of the ROV. 
\textbf{1)} The ROV initiates homing using acoustic positioning via USBL. 
\textbf{2)} Upon visual contact, it transitions to camera-based navigation using ArUco markers or manual navigation and data exchange through the optical modem. 
\textbf{3)} The ROV docks into the station, aligning with an inductive charger to charge and establish communication. 
\textbf{4)} After recharging or data transfer, it undocks and proceeds with the inspection mission.}
\label{fig:graphical_abstract}
\end{figure}

Resident, mobile, and persistent systems offer a promising solution to these limitations by enabling both continuous presence and spatial flexibility. Vasilijevic \cite{vasilijevic_portable_2024} introduced the concept of a portable seabed station, featuring a flexible instrumentation setup and an integrated mobile fly-out vehicle. In this concept, the vehicle initially navigates toward the station via long-range, low data-rate Ultra Short Baseline (USBL) communication. Upon establishing visual contact, it performs an autonomous docking manoeuvre guided by visual markers. At close range, it can also be remotely operated and exchange data through an optical modem. Once docked, it supports high-bandwidth communication and recharges via an inductive connector, as illustrated in Fig. \ref{fig:graphical_abstract}. 

Building on the shallow-water trials described in \cite{vasilijevic_portable_2024}, which relied entirely on manual operation, this study presents the first autonomous operational results of the infrastructure at both shallow and 90-meter depth. This includes the construction, deployment, and practical use of the system under realistic conditions. The results represent a significant step toward fully autonomous operation, particularly in the areas of docking and infrastructure inspection at depth, demonstrating the potential for persistent underwater presence. Additionally, procedures for acoustic homing were developed and validated in simulation.
\section{Methods}\label{Methods}
The operational concept at the core of this work is the deployment of a resident underwater vehicle capable of autonomous docking, recharging, communication, and inspection tasks. Realising this concept requires the integration of multiple subsystems into a coherent architecture that supports persistent and autonomous underwater presence.

These subsystems include the hardware components described in Section \ref{Hardware}, such as the physical infrastructure, vehicle configuration and sensor integration, and the modular design of the docking station.
Complementing the hardware, Section \ref{Software} outlines the simulation environment used for developing and testing autonomous behaviours.
Building on these foundations, Section \ref{Implementation} details the implementation and validation of key procedures that enable autonomous operation, specifically, visual guidance using ArUco markers, docking and inspection routines, and acoustic homing with a USBL system. These experiments provide practical insights into how such systems can be realised and deployed in real-world conditions.

\subsection{Hardware}\label{Hardware}
\subsubsection{Site Layout and Deployment}\label{Site_Layout}
The docking station was designed as a stand-alone multi-sensor platform, but both its physical design and software interfaces were developed to integrate with the existing infrastructure at the NTNU Oceanlab at 90\,m depth. 

Three main platforms are installed at the site: a large-class AUV/ROV docking station, which provides both power and network connectivity; an instrument rig that continuously collects environmental data such as salinity, water velocity, and turbidity; and a decommissioned pig loop platform, currently used as an observation object. 

The docking station used in this project was mounted on top of the pig loop platform. It is aligned using two guiding pins that interface with rails on the upper surface, ensuring both stability and a level mounting surface. Power and network interfaces are available on-site. For this use case, a 250\,W DC interface was used. 

The docking station was deployed with the resident ROV strapped to it, using R/V Gunnerus, a research vessel equipped with sufficient lifting capacity and a dynamic positioning system. 

\begin{figure}[!htbp]
\centering
\includegraphics[width=0.5\textwidth]{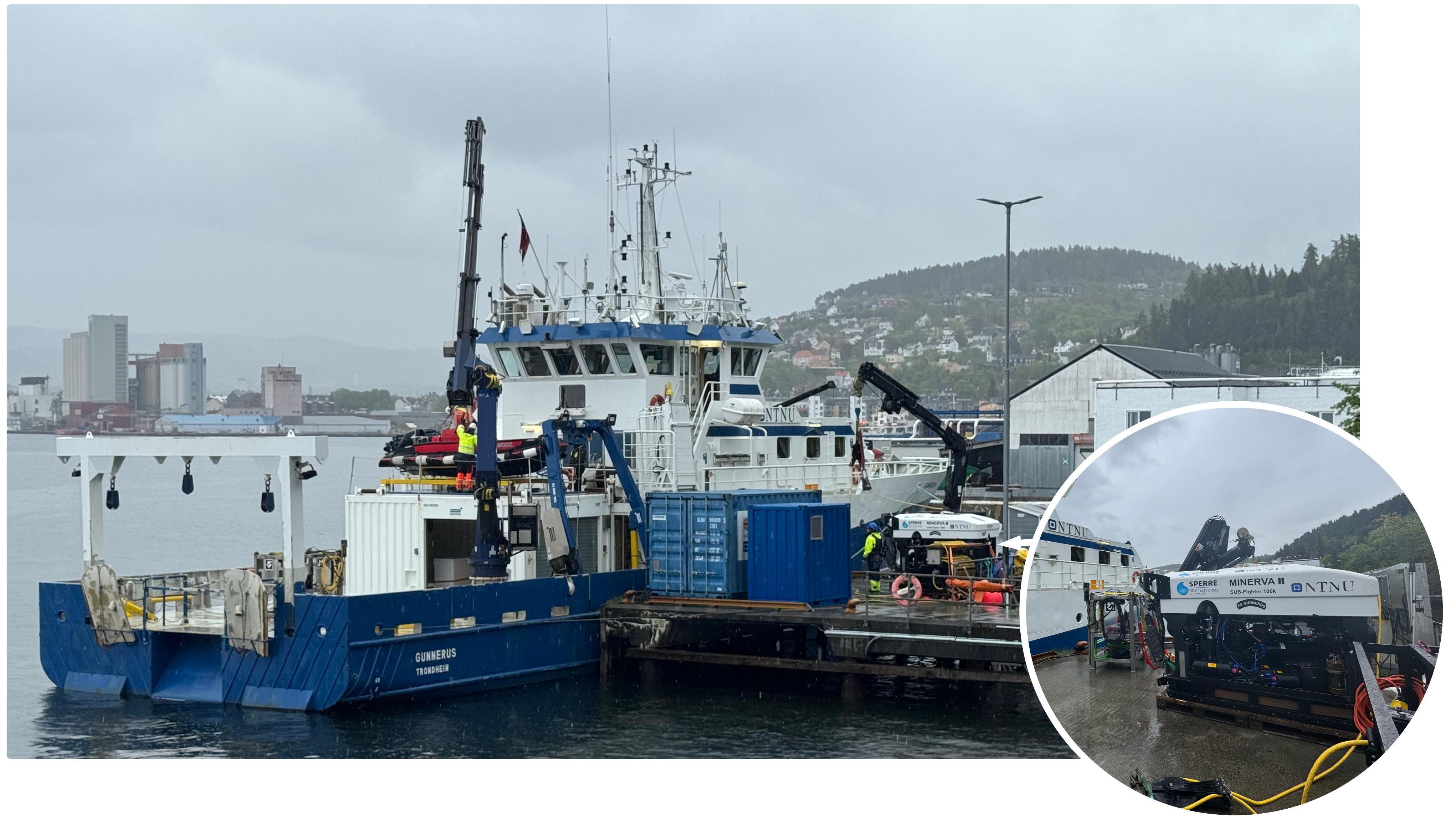}
\caption{R/V Gunnerus and the working-class ROV Minerva II during the deployment operation. The inset highlights Minerva II equipped with manipulators used to position and adjust the station underwater.}

\label{fig:Operational_Procedure}
\end{figure}

Deployment was further supported by our working-class ROV, Minerva II (both displayed in Fig. \ref{fig:Operational_Procedure}). The docking station was slowly lowered by the crane aboard R/V Gunnerus until it hovered just above the pig loop module. The ROV crew then guided the vessel into the correct position and used Minerva II’s manipulators to make the final adjustments.

\subsubsection{ROV}\label{ROV}
The ROV used in this project is the Blueye X3 \cite{blueye_x3}, a mini-class ROV weighing approximately 10\,kg, equipped with a forward-facing camera suitable for visual navigation, an onboard IMU, and a depth sensor. It also provides three external Guestports for additional sensors.

\begin{figure}[!htbp]
\centering
\includegraphics[width=0.5\textwidth]{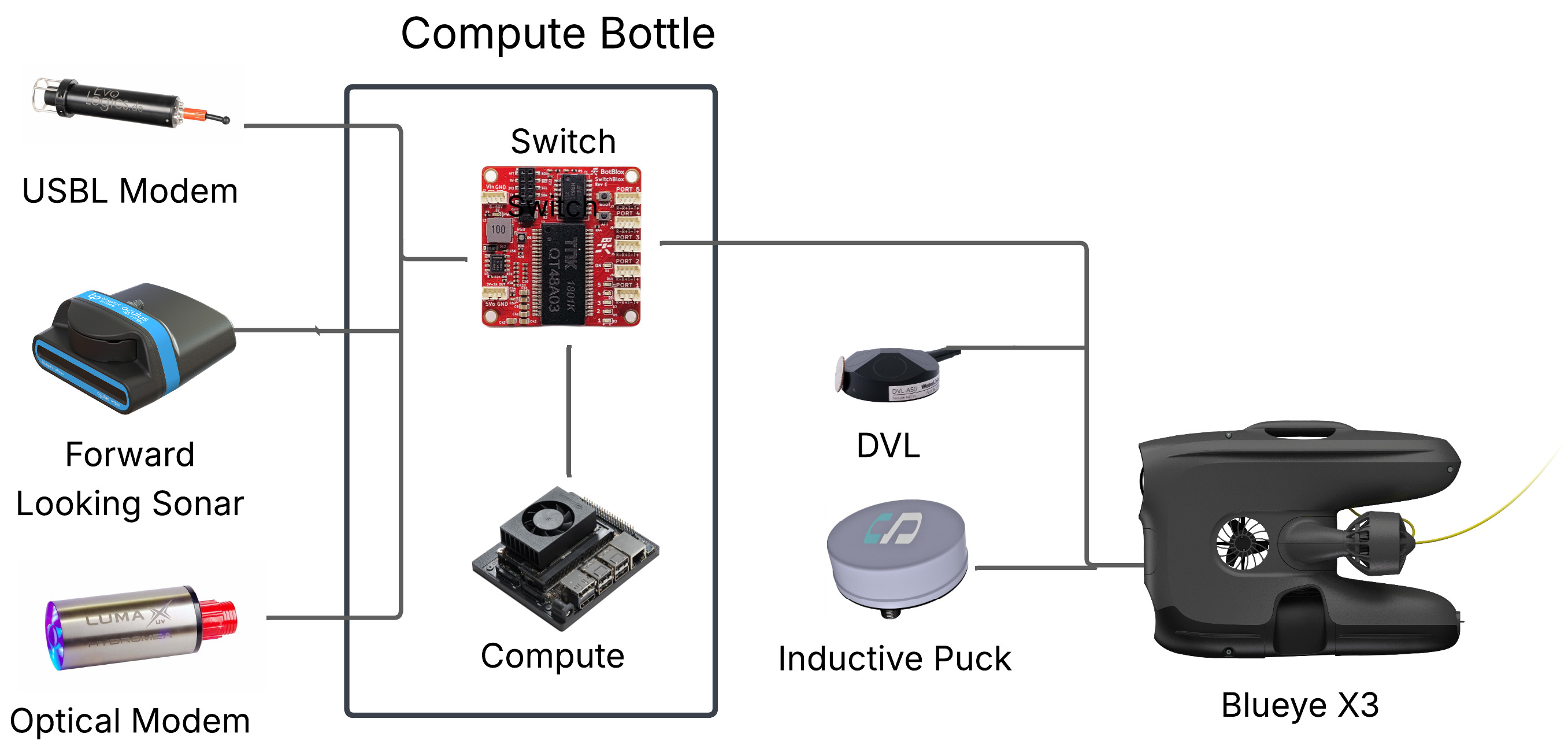}
\caption{Network layout of the sensors connected to the Blueye X3 ROV.}
\label{fig:network_layout}
\end{figure}

However, due to the complexity of our mission requirements, the available Guestports and onboard processing capacity were insufficient. To address this, an external processor bottle was added. The processor bottle connects to the Blueye via Ethernet and draws power directly from the vehicle’s battery. The complete network layout is shown in Fig. \ref{fig:network_layout}.
\newpage
The key components and their functions are listed below:
\begin{itemize}[leftmargin=*, label={}]
    \item \textbf{USBL Modem}: For long-range acoustic homing, the Evologics S2C R 18/34 USBL modem is mounted upward-facing on the rig, and a compact S2C T 18/34 modem is mounted downward-facing on the ROV\cite{Evologics_USBL}. 

    \item \textbf{Forward Looking Sonar}: An Oculus M750 multibeam sonar \cite{blueprint_oculus_m750d} provides real-time acoustic imagery for obstacle avoidance and environmental awareness.

    \item \textbf{Optical Modem}:  Optical communication is handled by Hydromea’s LumaX modem\cite{Hydromea_LUMA}, facilitating high-bandwidth, short-range optical communication, near the docking station or between vehicles.

    \item \textbf{DVL (Doppler Velocity Log)}: A Waterlinked A50 \cite{waterlinked_dvl_a50} directly interfaces with the Blueye via its dedicated port.

    \item \textbf{Inductive Charger}: Docking and power transfer are facilitated by the 250\,W Subsea Power Puck from Subsea USB\cite{BlueLogic_SPP}, which provides both power and high-speed data through inductive coupling.

    \item \textbf{Compute Unit}: 
    A Jetson Nano Developer Kit \cite{nvidia_jetson_nano_get_started} runs autonomy, perception, and communication processes. It is connected via an onboard switch and handles all sensor data aggregation and processing.
\end{itemize}

\subsubsection{Docking Station}\label{Docking_Station}
\begin{figure}[!htbp]
\centering
\includegraphics[width=0.5\textwidth]{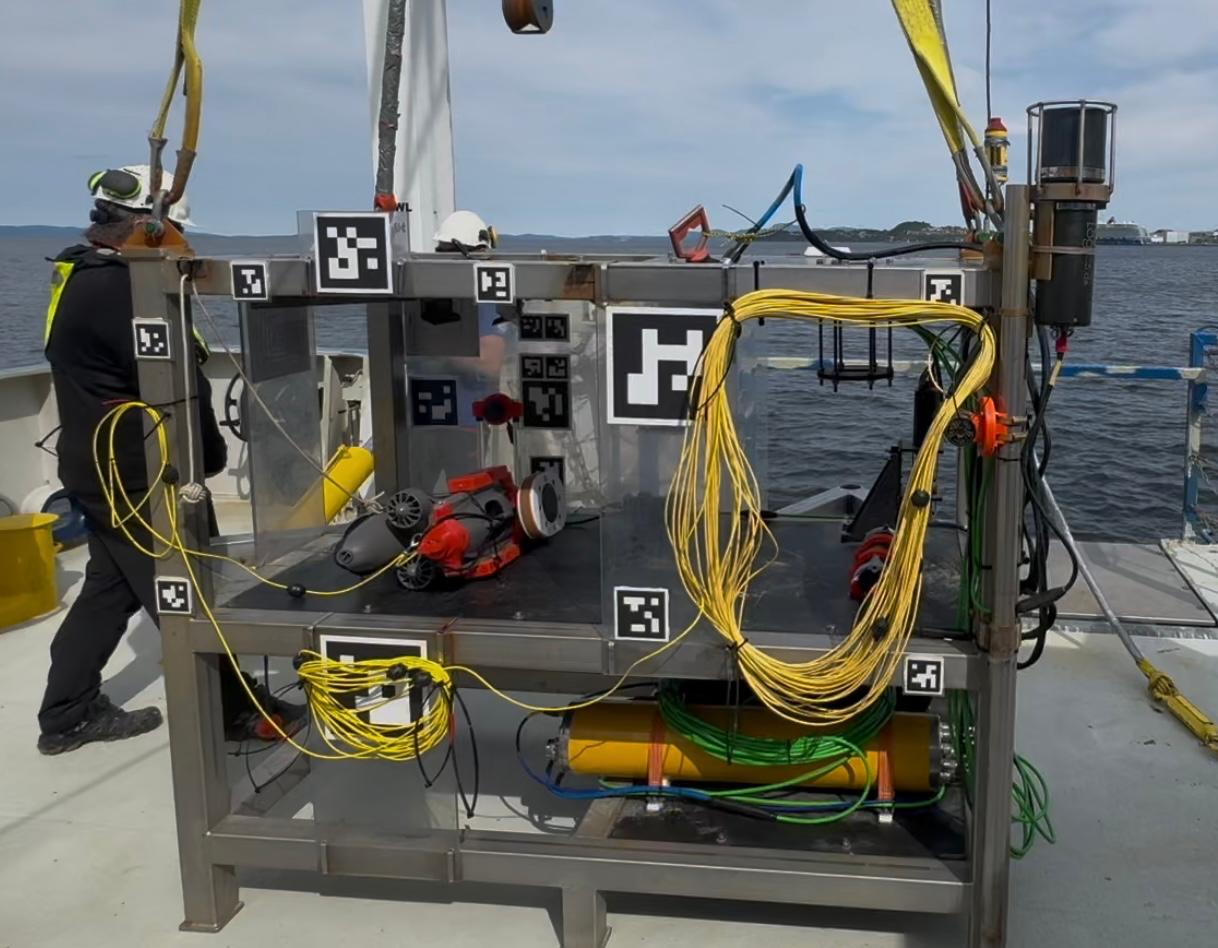}
\caption{Docking station aboard R/V Gunnerus prior to deployment. The ROV is secured in the main compartment, with a 20\,m tether below and an 80\,m reserve tether on the right. The USBL and optical modem are mounted on the right column.}
\label{fig:docking_station_on_ship}
\end{figure}

The docking station, shown in Fig. \ref{fig:docking_station_on_ship}, measures 2\,m in width and height, and 1\,m in depth. Constructed from stainless steel, it weighs approximately 400\,kg in air and 100\,kg when submerged. The structure is divided into two primary compartments: the lower compartment houses the data processing and network modules along with the battery bottle, while the upper compartment accommodates the ROV and various sensors, as well as the docking mechanism. The docking is facilitated by a 2D funnel-shaped guide on the top plate, which helps align the ROV's inductive charger with its counterpart on the station. A circular array of magnets then secures the vehicle in place. Additional instruments, such as the USBL receiver and optical modem, can be mounted on top or along the sides of the station, allowing for flexible sensor integration.

Given the complexity of resident operation at such depths and over extended durations, this study adopts a tethered configuration, allowing operational procedures to be separated into tethered and untethered experiments with the option for human intervention. In this setup, the ROV remains connected to the docking station via a data cable, enabling realistic trials in docking, inspection, and long-term deployment, while also permitting independent evaluation of navigation and homing algorithms using a second ROV.

To enable separate evaluation of surface-to-station homing and underwater inspection tasks, certain system resources were divided between two configurations. The resident ROV with an internal camera, IMU and depth sensor was equipped with an external DVL and the inductive charging puck. Further, during these resident experiments, the vehicle was operated in a tethered configuration, allowing closer observation, direct interaction, and rapid parameter adjustment during operation.

For homing approach experiments, a second ROV of the same type was configured with an external USBL Modem, a compute bottle, and a DVL to support acoustic positioning. This modular setup allowed independent testing of homing, guidance, and docking capabilities. This configuration provides the following communication channels:

\begin{itemize}[leftmargin=*, label={}]

    \item \textbf{Tethered:} Allows for Ethernet communication of up to 500\,Mbps, but no power transfer.

    \item \textbf{Inductive:} Enables 250\,W power and 100\,Mbps data transfer during docking; used for charging and log download.

    \item \textbf{Optical:} Achieves 4–10\,Mbps over 20–30\,m with good alignment; enables mid-bandwidth transfer without docking, though performance depends on water clarity.

    \item \textbf{Acoustic:} Offers reliable, long-range, low-bandwidth communication and positioning via a combined modem/USBL unit.
\end{itemize}

While the design by Vasilijevic \cite{vasilijevic_portable_2024} relies solely on battery power to maximise deployment flexibility, our prototype required a constant power connection to extend the operational window from hours to days. The station was deployed with an integrated battery bank. This was necessary to handle high power surges during both charging and startup. 

In the tethered ROV implementation, the Blueye ROV’s tether, operating via an Ethernet-over-powerline protocol, is converted to standard Ethernet and connected directly to the processor bottle. Rather than implementing a complex spooling system, a simpler approach was chosen: a 20-meter section of the tether is left free-floating in the water column with added buoyancy elements, while an additional 80 meters remains on reserve in case the ROV needs to be retrieved.

A further challenge specific to the Blueye X3 platform is that the robot can only be turned on and off by an external magnet, which needs to be placed near a reed switch located behind the LED array. To address this, an electromagnet was installed at the end of the docking funnel, controllable via the processor bottle, allowing remote activation or shutdown. For the current deployment, the ROV was continuously powered on to avoid the risk of it drifting away while turned off and unreachable. This underscores the importance of implementing a remote power control capability for fly-out vehicles.

\subsection{Software}\label{Software}
Development of navigation, guidance, and control at AURLab builds on a software-in-the-loop (SIL) approach, as described in \cite{waldum_virtual_2025}. Simulation enables virtual testing and development prior to field trials, allowing software errors to be identified and resolved early through SIL. The main development platform is a Gazebo-based simulator \cite{Koenig_Gazebo}, in which the behaviour of various platforms, including the vehicle discussed in this publication, can be simulated and corresponding control strategies developed.

A key aspect of this setup is that the simulation environment and control strategies for all ROV platforms used in the lab, including the mini-ROV, work-class ROV, and a medium-sized ROV under development, are built around a unified control architecture. The hydrodynamic parameters for the Minerva II and Blueye vehicles within the simulator were developed by Hoven \cite{holven2018control} and Alham \cite{alham2024enhancing}, respectively. This shared foundation ensures that algorithms developed and validated in simulation for one platform can be readily transferred to others with minimal modification.

\begin{figure}[!htbp]
    \centering
    \includegraphics[width=0.48\textwidth]{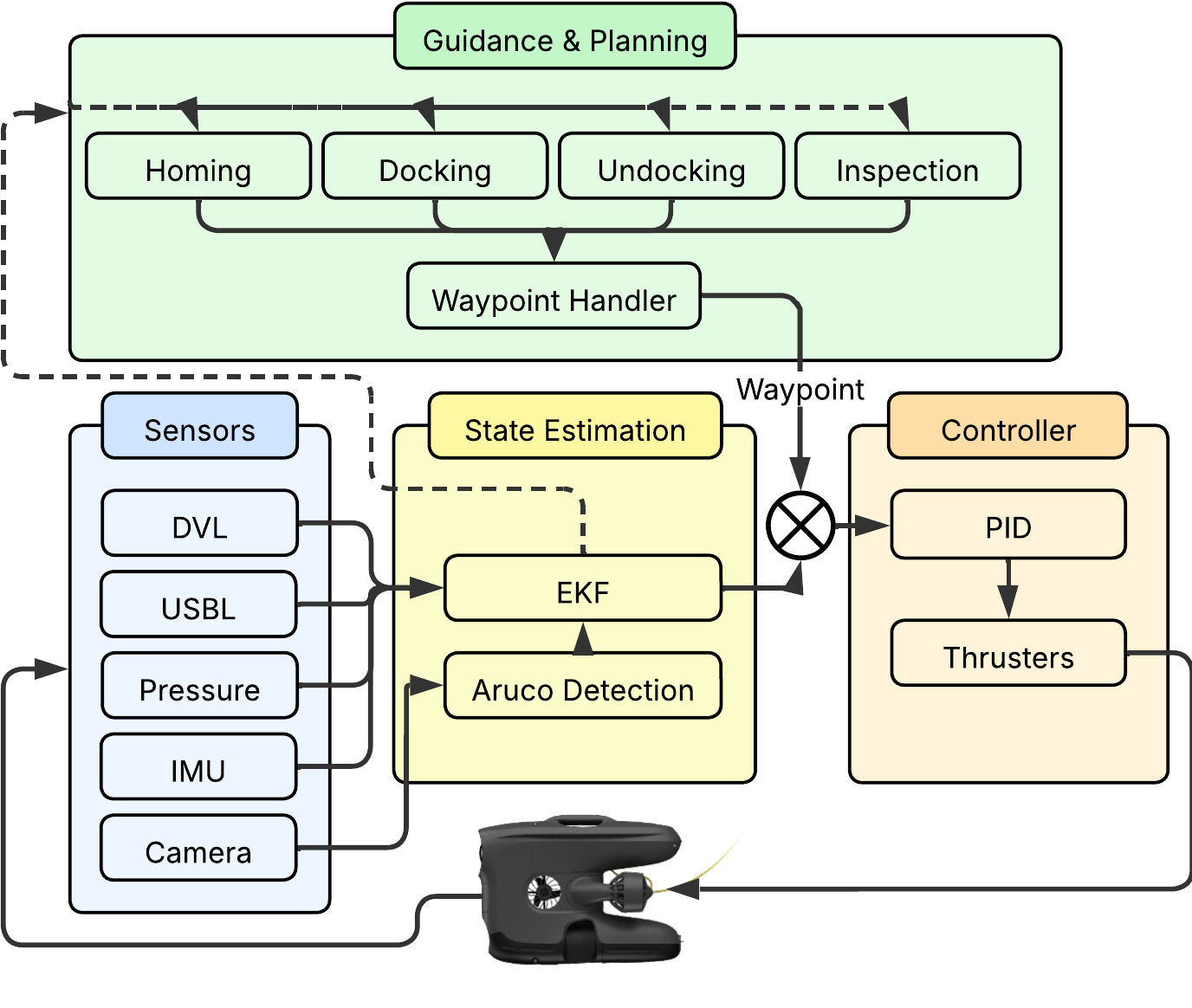}
    \caption{Software architecture for guidance, navigation, and control of the resident ROV.}
    \label{fig:software_architecture}
\end{figure}

To ensure a smooth transition from simulation to real-world deployment, the codebase is structured into three distinct components: simulation-specific code, hardware interface code, and control system code. The control components are shown in Figure \ref{fig:software_architecture}. The system is organized into sensing, state estimation, control, and a high-level guidance and planning modules. Sensor inputs, including DVL, USBL, pressure, IMU, and camera data, are fused within an Extended Kalman Filter (EKF), together with the visual pose estimates from ArUco marker detection. The resulting state estimate is used both for low-level control, where it is combined with waypoint references to generate control commands via a PID controller, and for high-level decision-making. The guidance and planning module implements behaviors such as homing, docking, undocking, and inspection, and adapts these behaviors based on feedback from the estimated system state.

\subsection{Implementation and Experiments}\label{Implementation}
The primary focus of this phase was to enable autonomous visual guidance, as described in Moslått \cite{moslatt2024guidance}. Visual markers were selected to support the docking procedure, and their optimal placement and size were determined through simulations of the station's general structure in the Gazebo simulator. This allowed the evaluation of marker visibility and ensured adequate coverage using the Blueye’s camera model, accounting for all relevant perspectives and approach angles to the docking station. The 3D models of the Blueye vehicle with the inductive charger mounted and the docking station prototype were obtained from Sudmann \cite{sudmann2023design}. To this, supplementary texture and colour were added, and an initial layout for the configuration of the tags was applied. ArUco markers were selected as the tag types due to their widespread adoption and well-recorded performance in underwater environments. OpenCV \cite{opencv_library} provides most of the required functionality through its well-documented ArUco library \cite{garrido_markers_nodate}. To determine the most suitable usage of the tags, a simulation experiment was conducted to explore the effects of bit pattern size and border thickness on detection performance; see Section \ref{ArUco_Experiments}.

Next, tag number, size and placement on the docking station were determined by investigating how tag size affects the maximum detection distance. For this, tag sizes 7\,cm, 15\,cm, and 25\,cm were tested in the simulator, and it was determined that a combination of different sizes provided the most robust performance during the docking procedure.

At least one large 22\,cm tag was placed on each of the four sides of the docking station to increase detectability from greater distances. However, as the vehicle approaches the docking station, these larger tags may no longer be fully visible in the camera frame and thus no longer provide position estimates. To compensate for this, smaller markers were placed strategically on the front side of the docking station, with a higher density around the entrance and inside the docking funnel areas where precise navigation is critical to avoid collisions and ensure proper alignment for docking.

\begin{figure}[!t]
\centering
\includegraphics[width=0.5\textwidth]{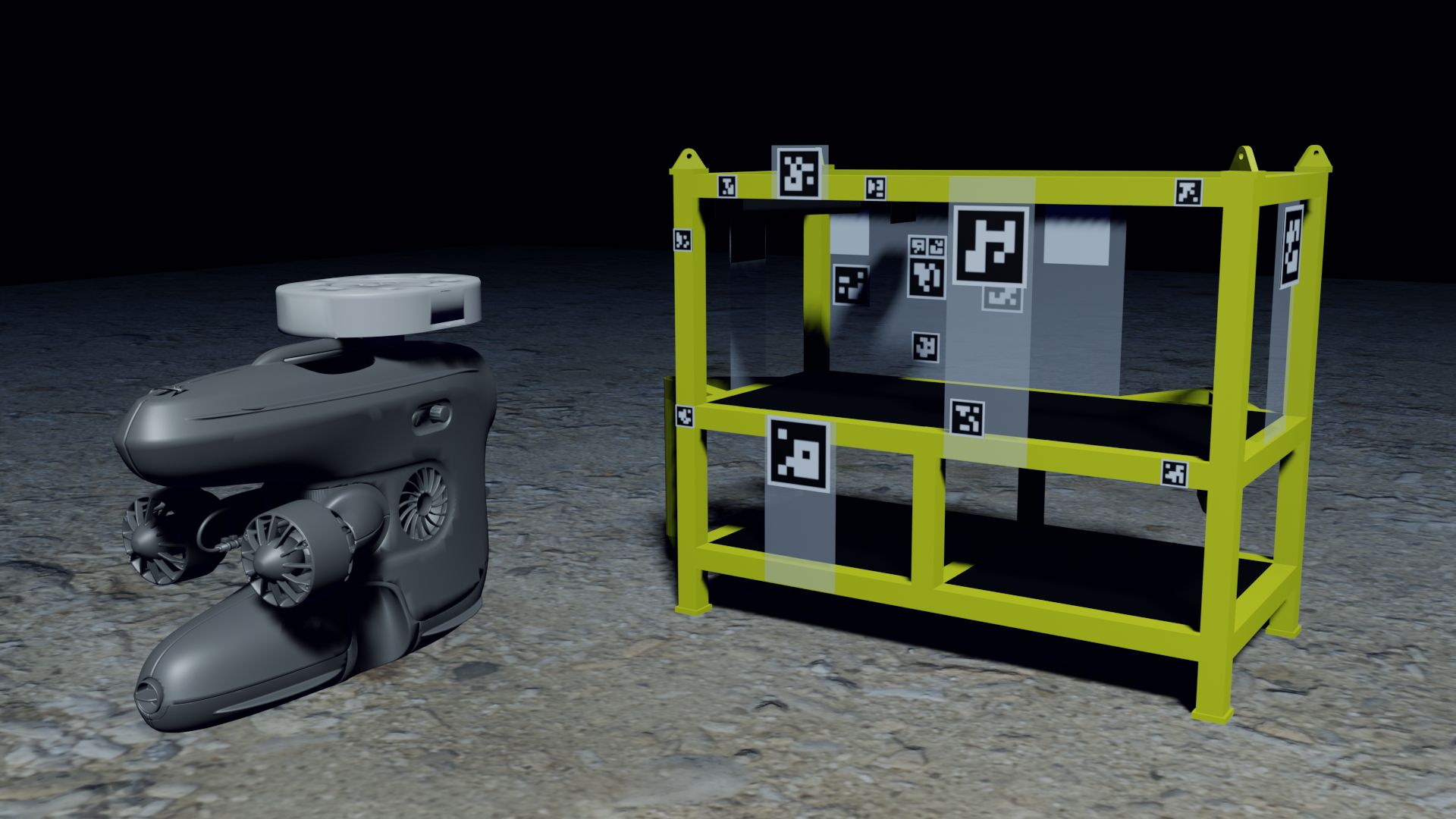}
\caption{The final docking station design viewed from the front as a Blender rendering, courtesy of \cite{moslatt2024guidance}.}
\label{final_design_actual_blender_right}
\end{figure}

The final design uses the first 21 tags from the aruco.DICT\_5X5\_50 dictionary and is illustrated in Fig. \ref{final_design_actual_blender_right}.

\begin{figure}[!t]
\centering
\includegraphics[width=0.5\textwidth]{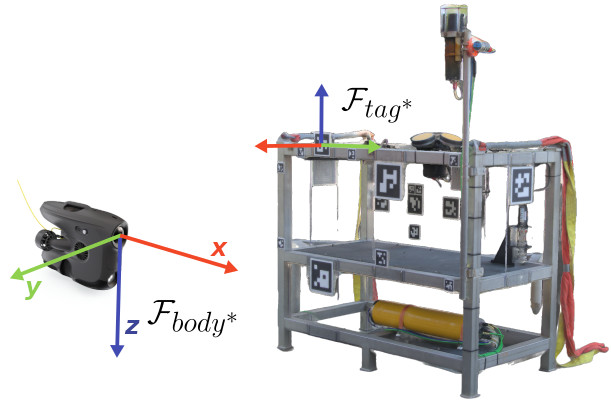}
\caption{Photogrammetry model of the docking station with coordinate frames of robot and docking station as defined by the center ArUco tag, courtesy of \cite{moslatt2024guidance}.}
\label{fig:photogram}
\end{figure}

The tag design was printed on thin aluminium plates with a matte finish to reduce glare and then mounted onto the docking station frame. Minor deviations between the digital model and the physical setup are expected during this mounting process. To account for these discrepancies, a photogrammetric model of the physical docking station with the ArUco tags was created using Metashape \cite{agisoft_metashape}. Using the software, the images were aligned by detecting shared feature points, from which a 3D point cloud was generated and used to construct a detailed 3D model. The resulting image network geometry is shown in Fig. \ref{fig:photogram}. Using manual annotation in Metashape, all tags were defined with their relative positions and orientations to the docking station.

\subsubsection{Docking Procedure}\label{Docking_Procedure}
Tag detection was implemented using the ROS 2 ArUco pose estimation pipeline \cite{aruco_ros2_docs}, building on the photogrammetric model described above. Within this pipeline, the image stream from the Blueye camera in Gazebo is converted to grayscale to reduce computational load. Marker detection is performed based on corner extraction and ID decoding, after which the robot pose is estimated relative to each detected tag. Given the known tag poses in the docking station reference frame, the robot’s global pose can then be recovered.

In addition to position estimation from visual cues provided by ArUco tags, the navigation system integrates data from the IMU and DVL using an EKF. This fusion improves the accuracy of position and velocity estimates while enhancing the robustness of the navigation system against individual sensor failures. If a sensor becomes unavailable, the system can still produce reliable state estimates by relying on the internal motion model and the remaining sensors. The EKF implementation used in this work is based on the ROS Robot Localisation package \cite{moore2016generalized}.

\begin{figure}[!t]
\centering
\includegraphics[width=0.5\textwidth]{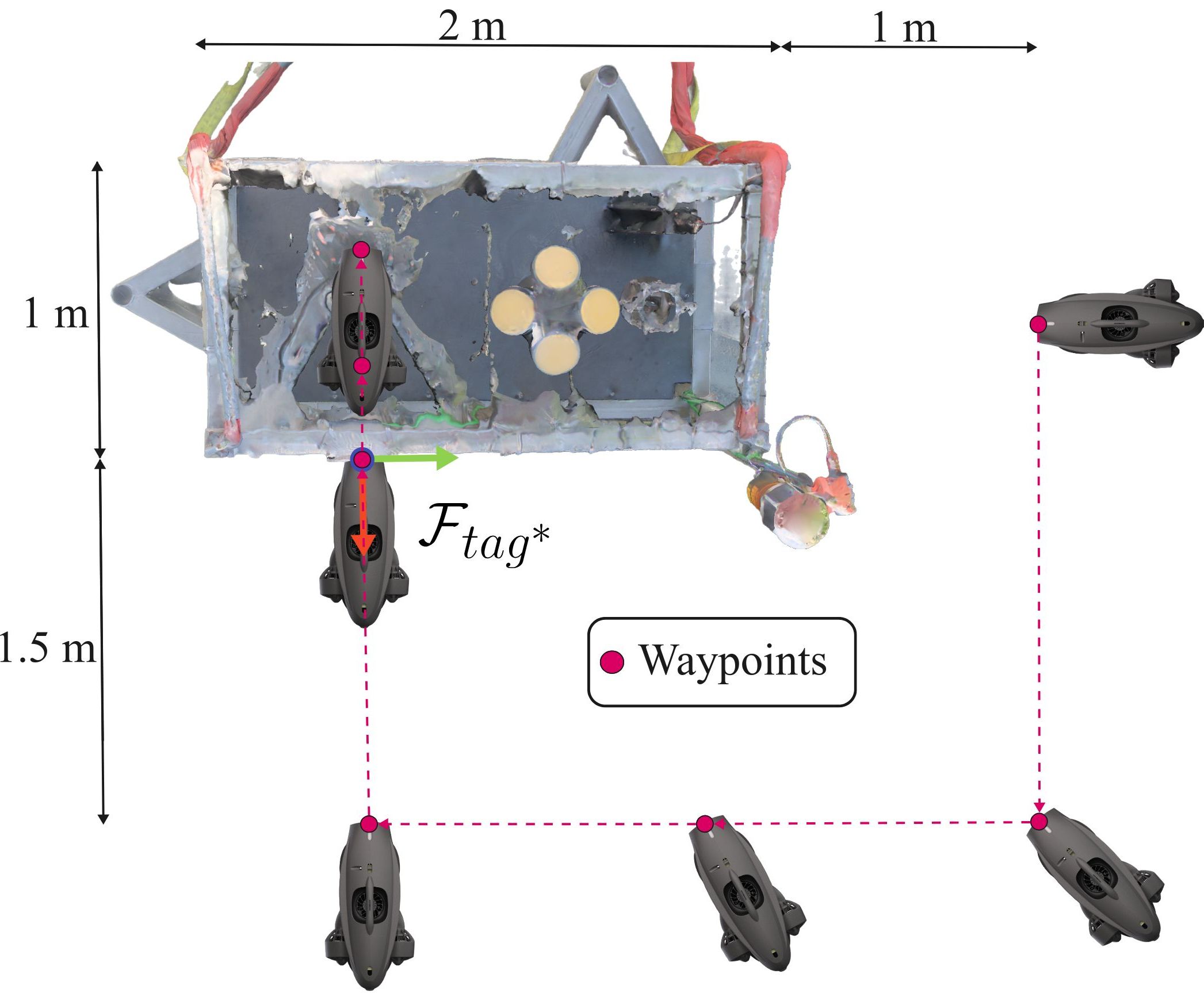}
\caption{Right-side loop waypoints, courtesy of \cite{moslatt2024guidance}.}
\label{fig:waypoint_right}
\end{figure}

The ROV state vector is 15-dimensional and includes the 6-DOF pose, velocities, and linear accelerations. The pose defines the position and orientation of the body frame relative to the reference tag frame $F_{tag^*}$, which corresponds to the primary marker located above the docking entry (Figs. \ref{fig:photogram} and \ref{fig:waypoint_right}). This tag defines the global reference frame, while the body frame $F_{body^*}$ follows the standard vehicle-fixed convention. Velocities and accelerations describe the motion of the body relative to the tag frame and are expressed in the body frame (see equation (\ref{eq:tag}) and (\ref{eq:x})).

\begin{equation}\label{eq:tag}
x := x_{\text{body}}^{\text{tag}^*}, \quad y := y_{\text{body}}^{\text{tag}^*}, \quad z := z_{\text{body}}^{\text{tag}^*}
\end{equation}

\begin{equation}\label{eq:x}
\bm{x} = \begin{bmatrix}
x, y, z, \phi, \theta, \psi, u, v, w, p, q, r, \dot{u}, \dot{v}, \dot{w}
\end{bmatrix}^{\mathrm{T}}
\end{equation}

The filter was tuned to place high confidence in the camera-derived pose estimates, allowing the EKF to closely track these measurements. A waypoint-based navigation scheme was developed to guide the ROV into the docking station, where it can connect to the inductive charging interface. Two U-shaped waypoint paths, referred to as the Left Loop and Right Loop (see Fig. \ref{fig:waypoint_right}), were defined to lead the vehicle into the docking station where it latches onto the magnetic array.

Each waypoint is defined by five parameters: its position in 3D space, a desired heading angle $\psi_d$, and a metric threshold representing the radius of an acceptance sphere. A waypoint is considered reached if the ROV enters this sphere while maintaining a heading within $\pm 5^\circ$ of the desired heading $\psi_d$.

When the autonomous docking sequence is initiated, the system selects the waypoint closest to the ROV as the initial target, $\textbf{WP}_{k+1}$, and assigns the ROV's estimated current position as the last reached waypoint, $\textbf{WP}_k$. Based on the selected target, the system determines whether the ROV will follow the Left Loop or the Right Loop. The waypoints are placed to optimise tag visibility, close enough in regards to detection accuracy and underwater attenuation, yet spaced to maximise the number of markers visible within the camera's field of view.

It is important to note that the docking algorithm described is too computationally demanding to run on the internal processor of the Blueye ROV. For testing, the algorithm was executed either on the surface computer used to pilot the ROV or, during autonomous trials, on the dedicated processing module housed in the external processor bottle.

\subsubsection{Inspection Procedure}\label{Inspection_Procedure}
Building on the established docking procedure, an autonomous inspection routine was implemented to visually survey all sides of the docking station. In this sequence, the ROV undocked and followed a trajectory around the right side of the station, returned to the center, and then completed a similar path on the left—covering the rear section of the rig in both instances before redocking. The inspection path was defined through a series of waypoints similar to those used for docking, with headings adjusted to ensure consistent camera coverage of structural and functional components. Each side of the routine lasted approximately 130 seconds.

\subsubsection{Acoustic Homing Procedure}\label{Acoustic_Homing_Procedure}
Before the ROV reaches the station, it must locate it using a USBL with modem functionality. The stationary modem initiates the exchange by pinging the smaller unit on the ROV; upon receiving a reply, it calculates the relative position offset in Northing and Easting and transmits this information back to the vehicle.

Due to the complexity of the acoustic water column at depth, the USBL system does not provide reliable depth measurements. However, since the depths of both the ROV and docking station are known, their vertical separation is directly calculated and incorporated into an EKF for waypoint estimation.

The ROV initially travels at the surface to a location approximately above the docking station. The upward-facing USBL modem, positioned at a depth of 90\,m, enables acoustic communication within a 100\,m surface diameter. Once the ROV receives three valid position fixes within 10 seconds, it begins navigating toward the USBL coordinate centre (with an intentional offset of -5\,m North and 0\,m East). Upon entering a defined target sphere around this position on the surface, the vehicle initiates its descent while maintaining a stable horizontal position.

The northward offset ensures that the docking station remains within the forward-facing field of view of the ROV’s cameras, allowing the vehicle to maintain a fixed heading toward north. Once the waypoint is reached, the ROV incrementally approaches the origin in steps of 0.5\,m while continuously checking for ArUco detections in the camera frame. When more than one tag is detected, the system transitions to docking mode. In this work, we limit ourselves to a simple analysis of the resulting behavior; for a more comprehensive explanation and elaborate reasoning, the reader is referred to Waldum et al. \cite{waldum2025docking}.



\section{Results}\label{Results}
The following section presents selected results from simulation, shallow-water experiments, and trials conducted at the 90\,m test site.

\subsection{Simulation Design Experiments}\label{ArUco_Experiments}

Preliminary design decisions regarding the placement and visual configuration of the ArUco markers were carried out in the Gazebo simulation environment. The objective was to establish an appropriate marker layout around the docking station by assessing geometric visibility and coverage under controlled visual conditions. Detailed optical modelling of the underwater environment was not included.

\begin{figure}[!htbp]
    \centering
    \includegraphics[width=0.4\textwidth]{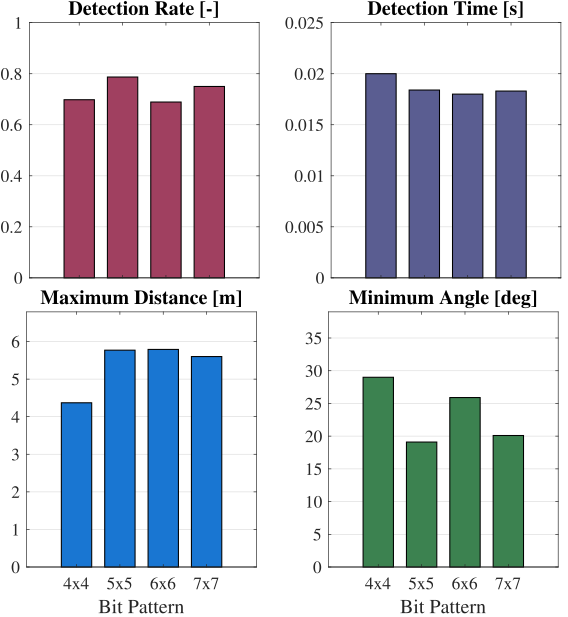}
    \caption{Comparison of detection performance metrics for tags with different bit patterns, courtesy of \cite{moslatt2024guidance}.}    
    \label{fig:bar_pattern}
\end{figure}

To evaluate the most relevant bit pattern configurations, comparative simulation experiments were conducted as shown in Fig. \ref{fig:bar_pattern}. In these experiments, a single ArUco tag with bit patterns ranging from 4×4 to 7×7 was assessed along the frontal approach path of the vehicle, while maintaining consistent positions and orientations across all iterations to ensure comparability. The results indicate that the 5×5 dictionary achieved the highest detection rate and one of the largest detection ranges, whereas detection latency remained similar across all tested configurations. The 4×4 pattern allowed detection at more oblique viewing angles but exhibited a reduced maximum range. Taking into account detection robustness, operational range, angular tolerance, and the requirement for a sufficient number of unique tag combinations, the 5×5 configuration was selected as a balanced solution for the final system.

\begin{figure}[htbp!]
    \centering
    \includegraphics[width=0.48\textwidth]{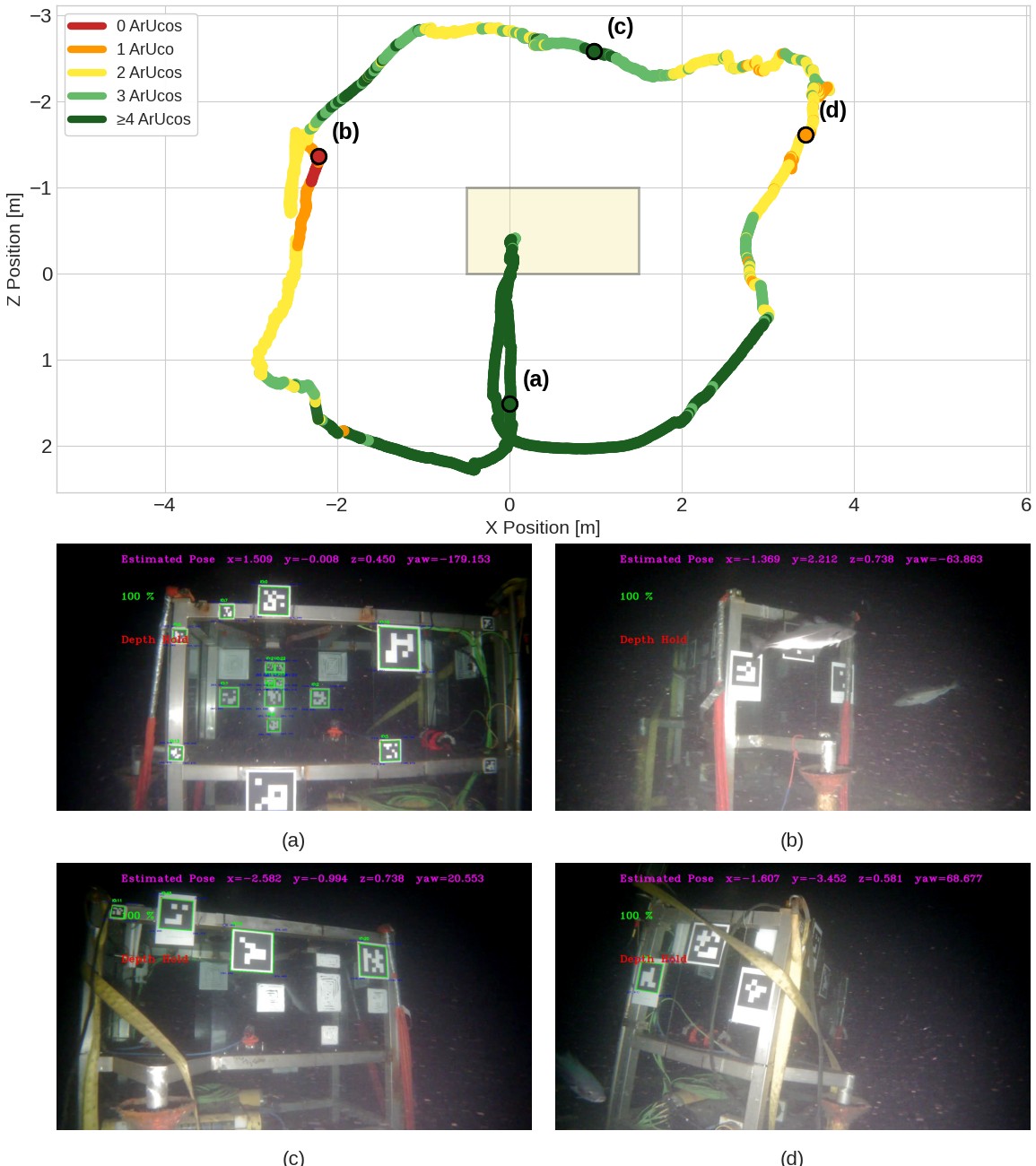}
    \caption{Manually flown robot trajectory colored by the number of detected ArUco tags during field deployment, with example camera frames illustrating detection performance near the docking station under real underwater conditions.}    
    \label{fig:Aruco_Detection_Figure_with_images}
\end{figure}

These preliminary simulations, together with empirical field experience, informed the final marker configuration, leading to the real-world detection results at the deep site shown in Fig. \ref{fig:Aruco_Detection_Figure_with_images}. The vehicle trajectory is colour-coded according to the number of detected ArUco tags. While exiting the station (a), up to 14 markers are detected simultaneously, resulting in the highest detection counts. Along the lateral and rear segments of the station (d and c), fewer markers are observed compared to the frontal view, although several remain visible in each frame. At the position indicated in (b), no marker is detected due to temporary occlusion by a fish crossing the camera’s field of view. In (c), the detection count is reduced as portions of the marker array are partially obstructed by equipment mounted on the station.

\subsection{Docking}
Preliminary tests at the Trondheim Biological Station (TBS) pier, conducted under good visibility and moderate currents, demonstrated high reliability as shown by \cite{moslatt2024guidance}. The docking algorithm successfully completed 13 consecutive front-facing dockings without failure. Additional trials from lateral offsets, where the ROV approached from 1–1.5\,m to the left and right, also resulted in successful dockings completed within 105 to 130\,seconds.

\begin{figure}[!htbp]
    \centering
    \includegraphics[width=0.5\textwidth]{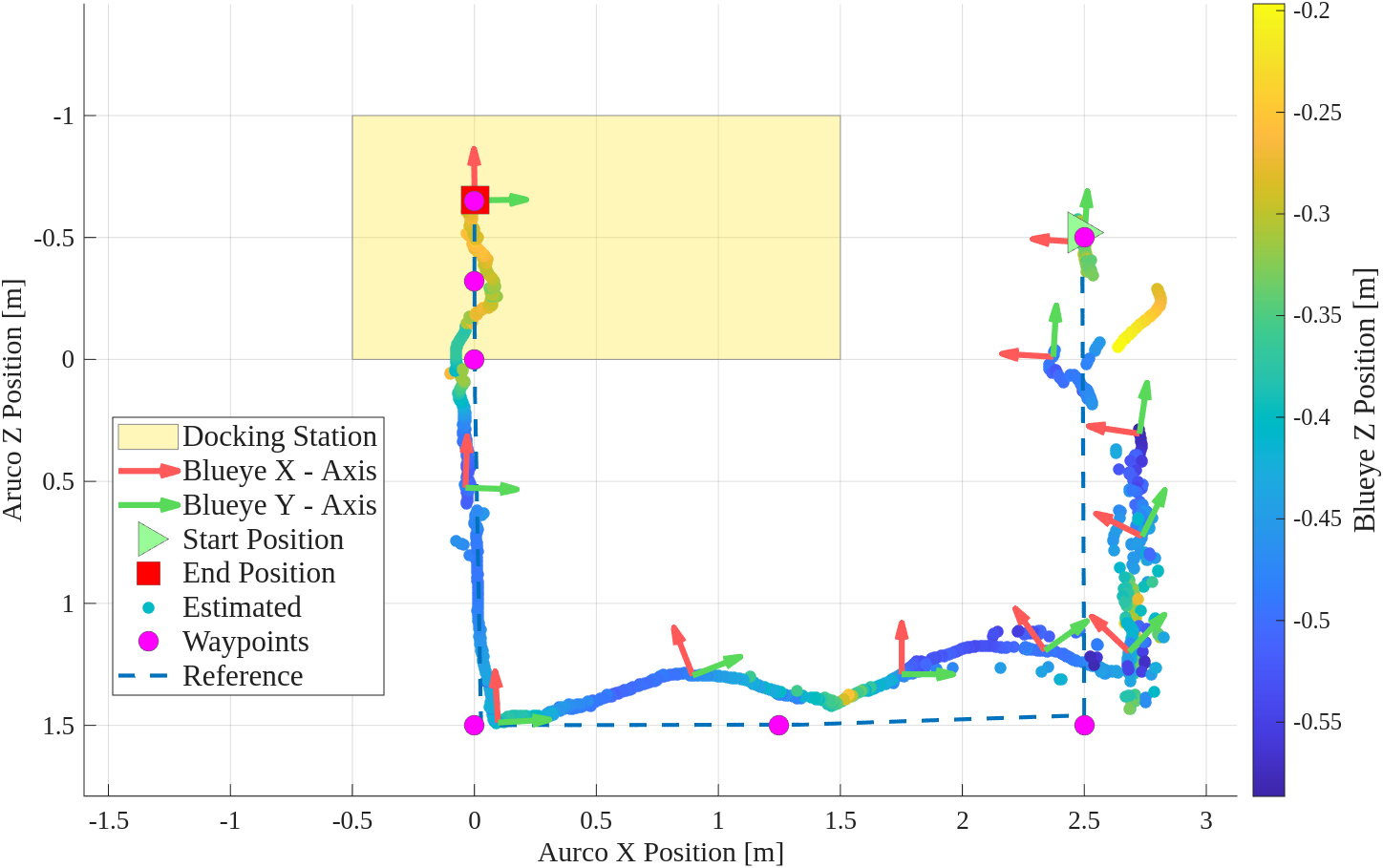}
    \caption{Docking sequence from the right-side loop at TBS. The vehicle begins at the green triangle and follows a sequence of waypoints (magenta circles) to reach the docking station. The estimated trajectory is shown in green, red and blue arrows denote the estimated vehicle X and Z axes and the green intensity the Y position. The dashed line represents the reference path, and the red square marks the final docking position. Slightly Modified from \cite{moslatt2024guidance}.}    
    \label{fig:right_side_result_TBS}
\end{figure}

An autonomous docking sequence from the right-side approach, conducted at TBS, is shown in Fig. \ref{fig:right_side_result_TBS}. The ROV begins approximately 2.5\,m to the right of the docking funnel and follows a U-shaped waypoint path designed to guide it around the station and align it for docking. Notably, the trajectory shows increased uncertainty at the beginning of the sequence. Based on operational experience, this occurs as the vehicle transitions from viewing a large ArUco tag to losing visual contact until it approaches close enough to detect the markers inside the station. A similar rise in pose uncertainty is visible around the corner point, where no markers are placed and the ROV must reorient itself before reacquiring visual references. The slight oscillations observed during the horizontal transverse are attributed to the ROV’s hydrodynamic behaviour, as it does not sway straight. During the final approach, there is strong alignment between the reference trajectory and the estimated position, up until the ROV reaches the inductive charger. At this point, it reduces speed and altitude, resulting in a slight positional offset.

\begin{figure}[!htbp]
    \centering
    \includegraphics[width=0.5\textwidth, trim=0 0 0 0, clip]{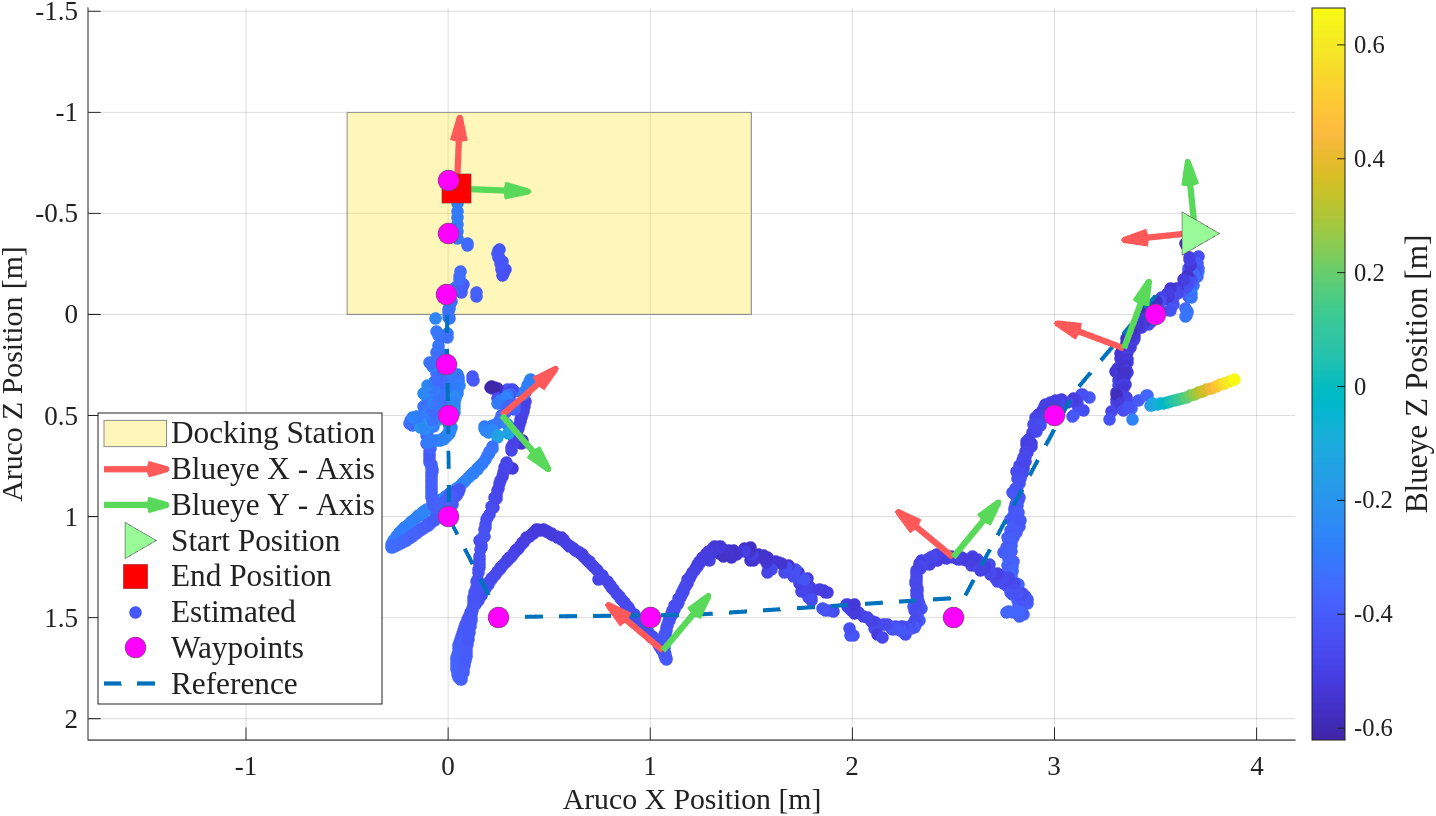}
    \caption{Docking sequence from the right-side loop at the 90\,msite. The vehicle begins at the green triangle and follows a sequence of waypoints (magenta circles) to reach the docking station. The estimated trajectory is shown in green, red and blue arrows denote the estimated vehicle X and Z axes and the green intensity represents the Y position. The dashed line represents the reference path, and the red square marks the final docking position.}    
    \label{fig:right_side_result_90m}
\end{figure}

The deep-docking site was less prone to environmental disturbances due to the absence of wave activity and the presence of highly predictable currents. The current proved to be a limiting factor during operations. By leveraging ADCP data, suitable operational windows were identified for docking and inspection tasks. Using the same waypoints as in the shallow-water tests, docking succeeded in 9 out of 10 trials from the front, 7 out of 10 trials from the left, and 6 out of 10 trials from the right. After adjusting the waypoints to maintain a larger stand-off distance (see Fig. \ref{fig:right_side_result_90m}), performance improved to 10 out of 10 trials from the left and 9 out of 10 trials from the right. The aborted approach during this trial was due to fish occluding the ArUco tags at the beginning of the approach. Docking duration was consistent across configurations, with an average time of approximately 140 seconds. A comprehensive analysis of the docking performance and parameters is provided in \cite{waldum2025docking}.

\subsection{Inspection}
\begin{figure}[!htbp]
    \centering
    \includegraphics[width=0.5\textwidth]{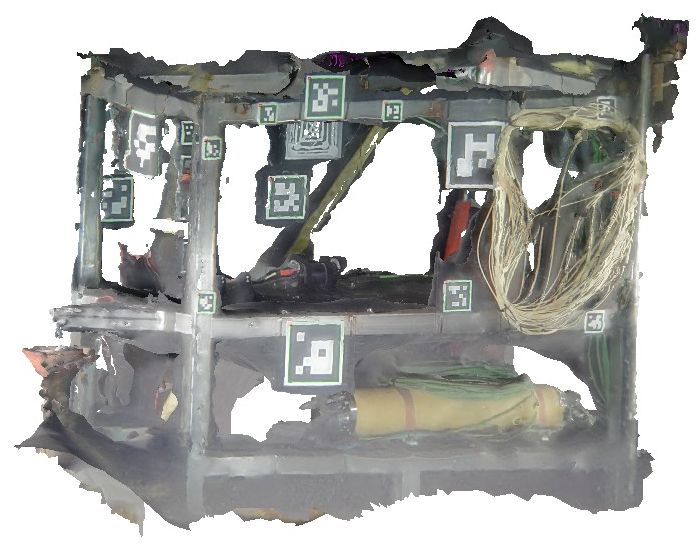}
    \caption{3D model created in Agisoft Metashape from inspection routine.}    
    \label{fig:inspection_photogrammetry}
\end{figure}

The entire station was successfully surveyed using the inspection algorithm. By combining the captured imagery with positional data, a 3D mesh was generated in Agisoft Metashape, as shown in Fig. \ref{fig:inspection_photogrammetry}.

\subsection{Acoustic Homing}
Due to equipment constraints, the acoustic homing trials were conducted using a Kongsberg MicroPAP system \cite{kongsberg_micropap} in combination with a Kongsberg C-Node transponder \cite{kongsberg_cnode}. A key limitation of this configuration is that the C-Node does not provide a direct interface to the ROV. To address this, the position estimates obtained from the bottom-side modem were forwarded to the guidance module. This workaround was feasible in the present setup because the ROV remained tethered, allowing access to the modem network. However, for future fully untethered deployments, a direct communication interface between the onboard vehicle and the acoustic positioning system will be essential.

Using this setup, a total of 340 USBL position updates were received over a 569.11\,s operation, corresponding to an effective update rate of 0.60\,Hz. To estimate the measurement standard deviation, a rolling window smoothing of the USBL trajectory was applied and visually verified. The deviation of the raw measurements from this smoothed trajectory resulted in a standard deviation of $0.089\,\mathrm{m}$. The system successfully guided the vehicle to the target location.

\section{Discussion}\label{Discussion}

\subsection{Docking}
When comparing the shallow and 90\,m trials in Fig. \ref{fig:right_side_result_TBS} and Fig. \ref{fig:right_side_result_90m}, it can be concluded that the sway motion is less stable at the 90\,msite and requires more frequent adjustments. This is likely due to the additional components mounted on the ROV, which alter its hydrodynamic behaviour. Additionally, Fig. \ref{fig:right_side_result_90m} shows two anomalies, one between the first and second waypoint and one toward the end of the trajectory. Both were caused by a temporary occlusion of one of the ArUco tags by a fish. On the shallow side during short-term visual occlusions, the control system should rely on compass-based heading estimates to maintain trajectory stability and ensure continuity of navigation. At the 90\,m side, however, significant magnetic drift, likely caused by interference from the metallic structure, persisted throughout the operation. As a result, the compass measurements were excluded from the EKF. In the upcoming deployment, this issue will be addressed by integrating a higher-grade IMU within the compute bottle to improve heading stability. Nevertheless, the ROV was able to recover once the tag became visible again, realign with the reference path, and complete the docking successfully.


One of the primary sources of error during the 90\,m-docking compared to the shallow water trials was reduced visibility at 90\,m depth due to the absence of external lighting. Limited field of view, particularly in terms of depth perception, became a critical constraint. This was especially evident during the left and right approach loops, where the robot previously detected the ArUco markers inside the station while navigating around the corner. These corner regions emerged as key points of failure. Further, several tags were covered by the rigging equipment, as shown in Fig. \ref{fig:Aruco_Detection_Figure_with_images} (b) and (d). However, increasing the stand-off distance improved reliability to 10/10 (left) and 9/10 (right) successful docking trials. 

\subsection{Latching}
We found the magnetic latching solution introduced by \cite{vasilijevic_portable_2024} to be highly reliable at the 90\,m site. The latching system consists of a circular array of permanent magnets embedded in the funnel head on the "roof" of the docking station, which mate with a matching pattern of corrosion-resistant steel bolts mounted around the inductive puck on the ROV. For a more detailed description, we refer the reader to \cite{vasilijevic_portable_2024}. 

Despite experiencing strong currents of up to 0.8\,m/s outside the station, the acrylic side walls appear to shield the ROV effectively from these disturbances. The magnetic alignment system holds the ROV securely in place during charging while still allowing deliberate undocking through applied thrust.

In contrast, previous tests conducted closer to the surface revealed that wave-induced forces can be strong enough to dislodge the ROV, even several meters below the waterline, causing it to drift out of the docking station.

\subsection{Inspection}
\begin{figure}[!htbp]
    \centering
    \includegraphics[width=0.5\textwidth]{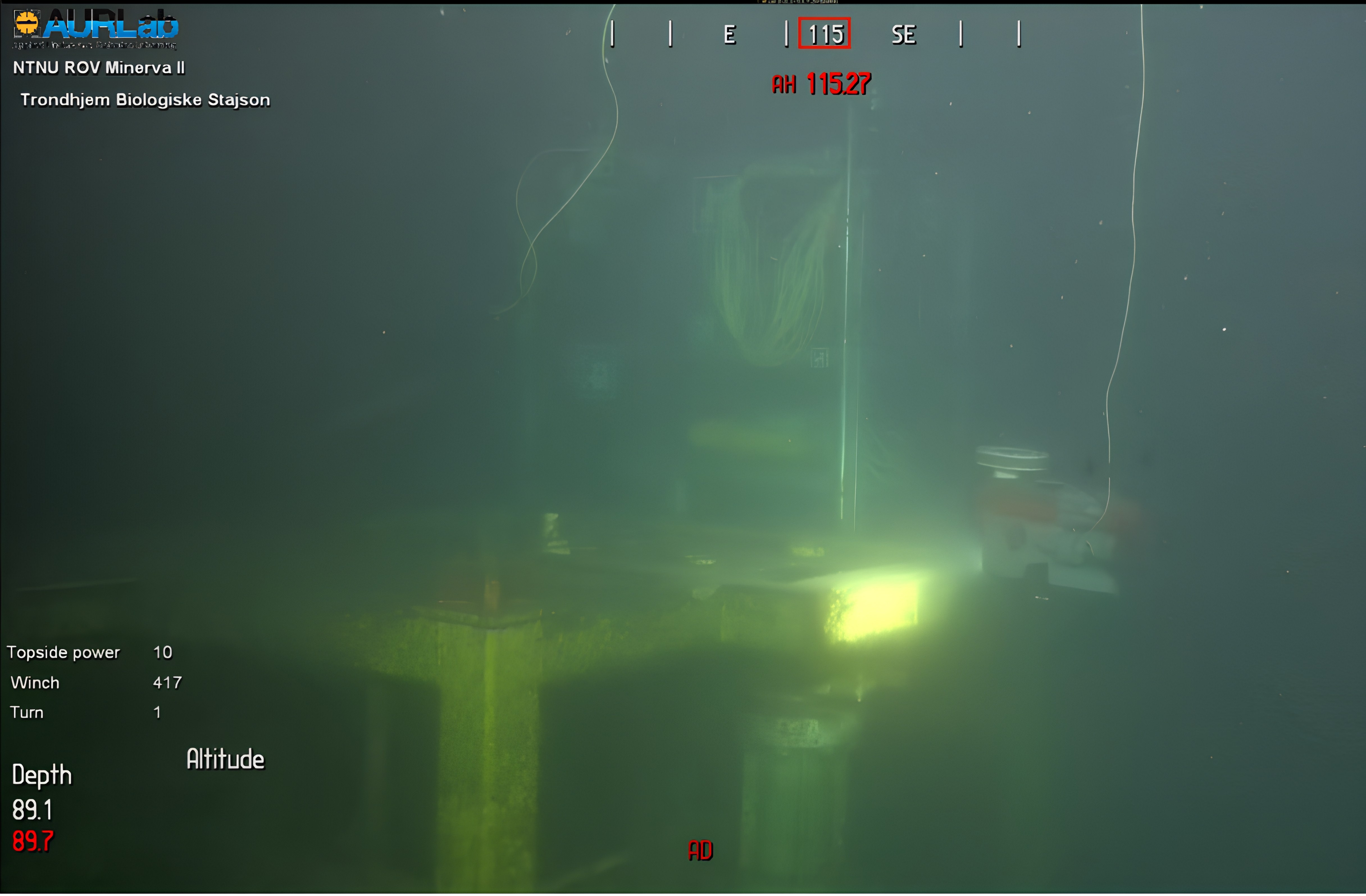}
    \caption{ROV during manual inspection of the pigloop module on which the docking station is mounted.}    
    \label{fig:photgrammetry_pigloop}
\end{figure}

The inspection task described in Section \ref{Inspection_Procedure} is effective as long as the surrounding structure provides sufficient optical tags to support visual navigation. However, as shown in Fig. \ref{fig:photgrammetry_pigloop}, the docking station is mounted atop an offshore pig loop platform. To enable concurrent inspection of both the station and the platform, a more advanced approach is required. One possibility is to develop an algorithm capable of navigating the complex geometry of the platform using a preliminary model generated from a prior manual inspection. The robot could then localise itself and plan its trajectory based on the recognition of similar structural features.

\subsection{Power and Size Restrictions}
Power management was a central consideration in the original system design. A soft-start procedure was implemented to limit inrush currents during startup, and a battery bank was integrated to buffer peak loads during charging, ensuring stable operation. With the inductive charger drawing up to 150\,W, the data processing bottle consuming around 80\,W at idle, and additional losses from long cable runs and DC transmission, continuous battery-powered operation was not feasible. These limitations were known and accepted trade-offs in favour of modularity and system reliability during early field trials conducted near existing infrastructure.

The large frame, likewise, was deliberately built to accommodate a wide range of sensors and experimental configurations, providing flexibility during the system’s initial development phase. However, in practical deployments, much of this capacity remained unused. As a result, the setup requires a large support vessel and heavy lifting equipment to deploy the docking station, despite the ROV itself being operable from shore. The main logistical constraint lies in lowering the structure to depth, not in the vehicle’s capabilities.

We are improving system logistics to enable battery-powered operation by developing a second, smaller, modular frame with a reduced sensor payload. 


\section{Conclusion}
In this work, we presented a multisensor docking station for a mini-class ROV, covering its conceptual design, the use of a digital twin for behaviour development, shallow-water testing, and final deployment. The system was operational for one month, with frequent use of the vehicle.

With the long-term goal of enabling fully autonomous flyout missions, we outlined the different phases of operation: acoustic homing, docking and undocking, and inspection. We described how each phase was tested using both resident and non-resident vehicles. Additionally, we detailed the software architecture, sensor configuration, and onboard compute used to support autonomous behaviours.

This infrastructure provides a valuable foundation for future applied research on autonomous subsea operations. We successfully demonstrated reliable and repeatable docking in both simulation and real-world environments, showed results for autonomous inspection of the docking station, and reported early trials of acoustic-based navigation. Operational insights were also shared, including suggestions for improving the mobility and deployment flexibility of such systems.

This work presented findings from the first stage of infrastructure deployment and laid the foundation for our next major milestone: the development of a fully autonomous resident vehicle. A key challenge ahead, both for this project and the broader field, is the creation of robust mission planning and fault management systems capable of handling short, complex missions in dynamic underwater environments.

\section*{Acknowledgment}
We gratefully acknowledge the support of our design team Dana Yerbolat, Elena Marie Kirchman, Jenny Krokstad, and Ai-Nhi Hoang, who supported us in building practical and applied solutions to mount all our sensors. We would like to thank Mahmoud Hussein Abdelrazik Hassan for his contributions to the sonar integration and experimental design. We also gratefully acknowledge Jakob Odenwald for developing a custom mounting frame for the Blueye ROV, which significantly supported and facilitated the practical aspects of our work. Finally, we extend our sincere thanks to the AURLab engineering team; this project would not have been technically or logistically possible without their expertise and support. 

This work was supported by the Research Council of Norway and carried out as part of the Safeguard and Centre for Autonomous Robotics Operations Subsea (CAROS) projects.

\bibliographystyle{ieeetr} 
\bibliography{references}  

@INPROCEEDINGS{Koenig_Gazebo,
  author={Koenig, N. and Howard, A.},
  booktitle={2004 IEEE/RSJ International Conference on Intelligent Robots and Systems (IROS) (IEEE Cat. No.04CH37566)}, 
  title={{Design and Use Paradigms for Gazebo, an Open-Source Multi-Robot Simulator}}, 
  year={2004},
  volume={3},
  number={},
  pages={2149-2154 vol.3},
  doi={10.1109/IROS.2004.1389727}}

@inproceedings{vasilijevic_portable_2024,
	address = {Singapore, Singapore},
	title = {Portable and {Flexible} {Seabed} {Station} with a {Fly}-{Out} {Vehicle}},
	copyright = {https://doi.org/10.15223/policy-029},
	isbn = {979-8-3503-6207-7},
	url = {https://ieeexplore.ieee.org/document/10682225/},
	doi = {10.1109/OCEANS51537.2024.10682225},
	language = {en},
	urldate = {2025-06-09},
	booktitle = {{OCEANS} 2024 - {Singapore}},
	publisher = {IEEE},
	author = {Vasilijevic, Antonio},
	month = apr,
	year = {2024},
	pages = {1--6},
}

@INPROCEEDINGS{waldum_virtual_2025,
  author={Waldum, Ambjørn Grimsrud and Fossdal, Markus and Basso, Erlend Andreas and Ludvigsen, Martin},
  booktitle={2025 IEEE Underwater Technology (UT)}, 
  title={{From Virtual Waters to Real Oceans: A Simulation-Driven Approach to ROV Control System Design}}, 
  year={2025},
  volume={},
  number={},
  pages={1-10},
  doi={10.1109/UT61067.2025.10947400}}

@misc{alham2024enhancing,
  author       = {Alham, R.A.A.},
  title        = {{Enhancing the Navigation and Fault Detection in small Remotely Operated Vehicles (ROVs)}},
  year         = {2024},
  howpublished = {\url{https://ntnuopen.ntnu.no/ntnu-xmlui/handle/11250/3155842}},
  type         = {Master's thesis},
  institution  = {NTNU Open}
}

@misc{holven2018control,
  author       = {Holven, E.B.},
  title        = {{Control System for ROV Minerva 2}},
  year         = {2018},
  howpublished = {\url{https://ntnuopen.ntnu.no/ntnu-xmlui/handle/11250/2564521}},
  note         = {Accessed: 16 June 2025},
  type         = {Master's thesis},
  institution  = {NTNU Open}
}

@mastersthesis{moslatt2024guidance,
  author       = {Bjørn-Magnus Moslått},
  title        = {{Guidance, Navigation and Control System for Autonomous Docking of Unmanned Underwater Vehicles}},
  school       = {Norwegian University of Science and Technology (NTNU), Department of Marine Technology},
  address      = {Trondheim, Norway},
  year         = {2024},
  type         = {Master's thesis},
}

@misc{sudmann2023design,
  author       = {Sudmann, S.},
  title        = {{Design and Prototyping of a Universal Docking Station for Small Underwater Vehicles}},
  year         = {2023},
  howpublished = {\url{https://ntnuopen.ntnu.no/ntnu-xmlui/handle/11250/3093883}},
  note         = {Accessed: 16 June 2025},
  institution  = {NTNU Open}
}

@article{opencv_library,
    author = {Bradski, G.},
    journal = {Dr. Dobb's Journal of Software Tools},
    title = {{The OpenCV Library}},
    year = {2000}
}

@misc{garrido_markers_nodate,
  author       = {Garrido, S. and Panov, A.},
  title        = {{Markers and Dictionaries, Detection of ArUco Markers}},
  year         = {n.d.},
  howpublished = {\url{https://docs.opencv.org}},
  note         = {Accessed: 16 June 2025}
}

@incollection{moore2016generalized,
  author    = {Moore, T. and Stouch, D.},
  title     = {{A Generalized Extended Kalman Filter Implementation for the Robot Operating System}},
  booktitle = {Intelligent Autonomous Systems 13},
  editor    = {Menegatti, E. and Michael, N. and Berns, K. and Yamaguchi, H.},
  publisher = {Springer International Publishing},
  address   = {Cham},
  year      = {2016},
  pages     = {335--348},
  doi = {10.1007/978-3-319-08338-4_25}
}

@misc{BlueLogic_SPP,
  title        = {{250W Subsea Power Puck (SPP)}},
  author       = {Subsea USB},
  howpublished = {\url{https://e-sea.bluelogic.no/main.aspx?page=article&artno=BB9867}},
  note         = {Accessed: 2025-06-30},
}

@misc{Hydromea_LUMA,
  title        = {{LUMA Underwater Optical Communication Modem}},
  author       = {Hydromea},
  howpublished = {\url{https://www.hydromea.com/luma-underwater-communication}},
  year         = {2025},
  note         = {Accessed: 2025-06-30},
}

@misc{Evologics_USBL,
  title        = {{EvoLogics S2C R USBL Positioning and Communication System}},
  author       = {EvoLogics},
  howpublished = {\url{https://www.evologics.com/usbl}},
  year         = {2025},
  note         = {Accessed: 2025-06-30},
}

@misc{nvidia_jetson_nano_get_started,
  author       = {{NVIDIA Corporation}},
  title        = {{Getting Started with Jetson Nano Developer Kit}},
  year         = {2024},
  howpublished = {\url{https://developer.nvidia.com/embedded/learn/get-started-jetson-nano-devkit}},
  note         = {Accessed: 2025-07-01}
}

@misc{agisoft_metashape,
  author       = {{Agisoft LLC}},
  title        = {{Agisoft Metashape}},
  year         = {2025},
  howpublished = {\url{https://www.agisoftmetashape.com/}},
  note         = {Accessed: 2025-07-01}
}

@misc{blueye_x3,
  author       = {{Blueye Robotics}},
  title        = {{Blueye X3 Underwater ROV}},
  year         = {2025},
  howpublished = {\url{https://www.blueyerobotics.com/rov/x3}},
  note         = {Accessed: 2025-07-01}
}

@article{nauert2023inspection,
  title={{Inspection and Maintenance of Industrial Infrastructure with Autonomous Underwater Robots}},
  author={Nauert, Franka and Kampmann, Peter},
  journal={Frontiers in Robotics and AI},
  volume={10},
  pages={1240276},
  year={2023},
  publisher={Frontiers Media SA},
  doi = {10.3389/frobt.2023.1240276}
}

@article{soldi2023monitoring,
  title={{Monitoring of Underwater Critical Infrastructures: The Nord Stream and Other Recent Case Studies}},
  author={Soldi, Giovanni and Gaglione, Domenico and Raponi, Simone and Forti, Nicola and d'Afflisio, Enrica and Kowalski, Paweł and Millefiori, Leonardo M. and Zissis, Dimitris and Braca, Paolo and Willett, Peter and Maguer, Alain and Carniel, Sandro and Sembenini, Giovanni and Warner, Catherine},
  journal={IEEE Aerospace and Electronic Systems Magazine}, 
  title={Monitoring of Critical Undersea Infrastructures: The Nord Stream and Other Recent Case Studies}, 
  year={2023},
  volume={38},
  number={10},
  pages={4-24},
  doi={10.1109/MAES.2023.3285075}}

@article{whitt2020future,
  title={{Future Vision for Autonomous Ocean Observations}},
  author={Whitt, Christopher and Pearlman, Jay and Polagye, Brian and Caimi, Frank and Muller-Karger, Frank and Copping, Andrea and Spence, Heather and Madhusudhana, Shyam and Kirkwood, William and Grosjean, Ludovic and others},
  journal={Frontiers in Marine Science},
  volume={7},
  pages={697},
  year={2020},
  publisher={Frontiers Media SA},
  doi = {10.3389/fmars.2020.00697}
}

@article{scholaert2020blue,
  title={{The Blue Economy: Overview and EU Policy Framework}},
  author={Scholaert, Frederikp and Margaras, V and Pape, M and Wilson, A and Kloecker, CA},
  journal={European Parliament},
  year={2020}
}

@misc{aruco_ros2_docs,
  author       = {Rafael Muñoz Salinas and Bence Magyar},
  title        = {{Aruco ROS 2 Documentation}},
  year         = {2011--2025},
  howpublished = {\url{https://aruco-ros2.readthedocs.io/en/latest/index.html}},
  note         = {Accessed: 2025-07-14},
}

@misc{waterlinked_dvl_a50,
  author       = {Water Linked},
  title        = {{DVL A50 Doppler Velocity Log}},
  howpublished = {\url{https://waterlinked.com/shop/dvl-a50-1248}},
  year         = {2025},
  note         = {Accessed: 2025-07-14},
}

@misc{blueprint_oculus_m750d,
  author       = {{Blueprint Subsea}},
  title        = {{Oculus M750d Dual‑Frequency Multibeam Sonar}},
  howpublished = {\url{https://www.blueprintsubsea.com/oculus/oculus-m-series}},
  year         = {2025},
}

@unpublished{waldum2025docking,
  author       = {Ambjørn Waldum and Gabrielė Kasparavičiūtė and Leonard Florian Tom Günzel and Bjørn-Magnus Moslått and Oscar Pizarro and Martin Ludvigsen},
  title        = {{Toward Reliable Underwater Docking in Dynamic Environments}},
  year         = {2026},
  note         = {Manuscript submitted for publication to IEEE Journal of Oceanic Engineering}
}

@misc{kongsberg_micropap,
  author       = {{Kongsberg Discovery}},
  title        = {microPAP Positioning System},
  url          = {https://www.kongsberg.com/discovery/navigation-positioning/micropap/micropap/},
  note         = {Accessed: 2026-03-25}
}

@misc{kongsberg_cnode,
  author       = {{Kongsberg Discovery}},
  title        = {cNODE Transponder},
  url          = {https://www.kongsberg.com/discovery/navigation-positioning/cnode-transponder/},
  note         = {Accessed: 2026-03-25}
}

\end{document}